\documentclass[lettersize,journal,twoside]{IEEEtran}
\usepackage[utf8]{inputenc}
\usepackage{dtk-logos} % for BibTeX stylized logo
\usepackage{graphicx}
\usepackage{hyperref}
\usepackage{threeparttable} % for table notes
\usepackage[dvipsnames]{xcolor} 
\usepackage{enumitem}
\usepackage{amsmath,amsfonts,amssymb,amsthm}
\hypersetup{colorlinks=true,allcolors=blue,linktocpage}
\usepackage{mathtools}

\usepackage{bm,upgreek}
\usepackage{nomencl}
\usepackage{multirow}
\usepackage{float}
\usepackage{subcaption}
\usepackage{array}
\usepackage[ruled,vlined,boxed]{algorithm2e}
\usepackage{algpseudocode}
\usepackage{booktabs}
\usepackage{pifont}
\usepackage{adjustbox}
\usepackage{siunitx}
\newcommand{\cmark}{\ding{51}} % checkmark symbol
\newcommand{\trace}{\mathop{\mathrm{Tr}}\nolimits}

\usepackage{orcidlink}
\newcommand{\ket}[1]{\lvert #1 \rangle}
\newcommand{\bra}[1]{\langle #1 \rvert}
\DeclareMathOperator*{\argmin}{arg\,min}
\algtext*{EndWhile}
\algtext*{EndFor}
\algtext*{EndIf}
\algtext*{EndFunction}
\algnewcommand{\SIf}[1]{\State\algorithmicif\ #1\ \algorithmicthen}
\algnewcommand{\SElseIf}[1]{\State\algorithmicelse\ \algorithmicif\ #1\ \algorithmicthen}
\algnewcommand{\SElse}{\State\algorithmicelse\ }
\algnewcommand{\SWhile}[1]{\State\algorithmicwhile\ #1\ \algorithmicdo}
\algnewcommand{\SFor}[1]{\State\algorithmicfor\ #1\ \algorithmicdo}
\algnewcommand{\SForAll}[1]{\State\algorithmicforall\ #1\ \algorithmicdo}
\makenomenclature

\IEEEoverridecommandlockouts

\begin{document}
	\title{Quantum Machine Learning and Grover's Algorithm for Quantum Optimization of Robotic Manipulators}
	
	\author{% 
		Hassen Nigatu\raisebox{0.5ex}{\orcidlink{0000-0002-4656-2725}}~\IEEEmembership{Member, IEEE},\,
		Gaokun Shi\raisebox{0.5ex}{\orcidlink{0009-0007-6884-3185}},~\IEEEmembership{Member, IEEE},\,
		Jituo Li \raisebox{0.5ex}{\orcidlink{0000-0003-1343-5305}},~\IEEEmembership{Member, IEEE},\,
		Jin Wang\raisebox{0.5ex}{\orcidlink{0000-0003-3106-021X}},~\IEEEmembership{Member, IEEE},\,
		Guodong Lu\raisebox{0.5ex}{\orcidlink{0000-0002-2762-9912}},~\IEEEmembership{Member, IEEE},\,
		and Howard Li\raisebox{0.5ex}{\orcidlink{0000-0002-6359-0349}},~\IEEEmembership{Senior member, IEEE},\,  
		\thanks{Manuscript received: January 12, 2025 ; Revised: June 6, 2025 ; Accepted: October 11, 2025.
			This paper was recommended for publication by Editor Cosimo Della Santina upon evaluation of the Associate Editor and Reviewers' comments.
			\par
			This research was supported by the Robotics Research Center of Yuyao (Grant No. KZ22308), the "Design and Fabrication of an Immersive Interactive VR Robot Controller" project under the Ningbo Yongjiang Talent Program (Grant No. Z22501), and the Zhejiang Talents Program. Additionally, it was supported by the National Natural Science Foundation of China (Grant No. 52275276).
			Hassen Nigatu is with the Robotics Institute of Zhejiang University (Yuyao Robotics Research Center), Yuyao Technology Innovation Center, No.\,479 Ye Shan Road, Yuyao, Ningbo Shi, Zhejiang Province, China. {Email: \tt\small hassen@ust.ac.kr}.
			Gaokun Shi, Jituo Li, Jin Wang, and Guodong Lu are with State Key Laboratory of Fluid Power and Mechatronic Systems, Zhejiang University, with Institute of Design Engineering, Zhejiang University, Hangzhou 310007, China and Robotics Institute of Zhejiang University, Yuyao 315400, China.
			Howard Li is with the department of Electrical and Computer Engineering, University of New Brunswick, Fredericton, NB, Canada.
			Correspondence: Howard Li, {\tt\small howard@unb.ca}.
			\par
			Digital Object Identifier (DOI): see top of this page.}	
	}	
	
	\maketitle
	
	\begin{abstract}
		Optimizing high-degree-of-freedom robotic manipulators requires searching complex, high-dimensional configuration spaces, a task that is computationally challenging for classical methods. This paper introduces a quantum-native framework that integrates Quantum Machine Learning (QML) with Grover's algorithm to solve kinematic optimization problems efficiently. A parameterized quantum circuit is trained to approximate the forward kinematics model, which then constructs an oracle to identify optimal configurations. Grover's algorithm leverages this oracle to provide a quadratic reduction in search complexity. Demonstrated on {simulated} 1-DoF, 2-DoF, and dual-arm manipulator tasks, the method achieves significant speedups—up to 93x over classical optimizers like Nelder-Mead—as problem dimensionality increases. This work establishes a foundational, quantum-native framework for robot kinematic optimization, effectively bridging quantum computing and robotics problems.
	\end{abstract}
	
	%%%%%%%%%%%%%%%%%%%%%%%%%%%%%%%%%%%%%%%%%%%%%%%%%
	\section{Introduction} \label{sec:Intro}
	%%%%%%%%%%%%%%%%%%%%%%%%%%%%%%%%%%%%%%%%%%%%%%%%%
	
	\IEEEPARstart{R}{obotic} manipulators are pivotal in modern automation, manufacturing, and precision tasks across industries including automotive assembly, medical surgery, and aerospace \cite{Ghafil2020,Hazarika2018,Siciliano2016}. These systems require precise end-effector positioning, which is achieved through inverse kinematics (IK) to determine the joint configurations corresponding to the target poses \cite{Tung2024,Nigatu2023}. Classical IK optimization methods perform adequately for low-DoF manipulators with simple workspaces but encounter limitations in high-dimensional, nonlinear manifolds \cite{Watterson2020,Park2017}. In parallel robot optimization or design candidate selection, challenges intensify due to nonlinear constraints, high-dimensional spaces, and mixed discrete-continuous parameters \cite{Dai2020}. Exhaustive searches over discretized joints scale exponentially as $O(k^N)$ for $N$-DoF systems with $k$  
	\begin{figure}[h!]
		\centering
		\includegraphics[width=0.85\columnwidth]{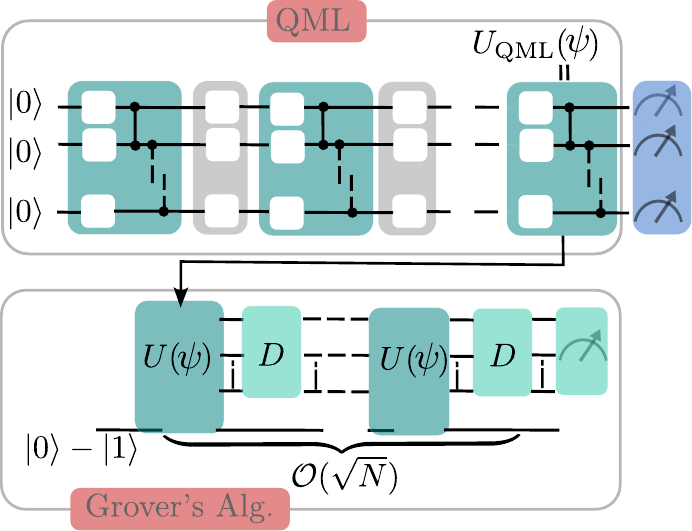}
		\caption{Integration of QML and Grover's algorithm. The QML block uses parameterized gates $U_{\text{QML}}(\psi)$ to encode learned kinematic patterns, biasing the initial state. This feeds into Grover's, where the oracle $U(\psi)$ marks solutions, and the diffusion operator $D$ amplifies them over $\mathcal{O}(\sqrt{N})$ iterations. Measurement yields optimal configurations with high probability.}
		\label{fig:integration}
	\end{figure}	
	values per joint, while advanced algorithms exhibit polynomial scaling, becoming intractable for high-DoF manipulators.
	
	This computational bottleneck motivates quantum computing, leveraging superposition and entanglement for parallel processing and asymptotic advantages in search and optimization \cite{Abbas2024,Havlicek2019,Jerbi2023}. Grover's algorithm searches unsorted databases of $M$ elements in $O(\sqrt{M})$ queries, offering quadratic speedup over classical $O(M)$ methods \cite{Grover1996}. In high-DoF robot IK optimization where $M \sim k^N$, this enables efficient exploration of constrained configuration spaces. Quantum approaches handle exponential complexity in non-convex landscapes, advancing robotics beyond classical limits as hardware matures. Consequently, quantum computing applications in robotics are expanding across multiple fronts, including Grover's algorithm for optimal control \cite{electronics14132503}, quantum annealing for IK via QUBO formulation with 30x speedups \cite{salloum2025annealing}, QML for robot test oracles \cite{wang2025qml}, quantum-inspired reinforcement learning for control \cite{yu2024quantum}, Grover-based classification \cite{du2021grover}, quantum-inspired sliding-mode control \cite{fazilat2025qsmc}, and quantum deep RL for navigation \cite{kirchner2024qdrl}.
	
	{However, existing approaches exhibit fragmented quantum-classical pipelines. For instance, while \cite{electronics14132503} uses Grover's search standalone and \cite{salloum2025annealing} relies on specialized annealing hardware, none integrate QML for learning with Grover's search within a unified quantum-native framework—an end-to-end pipeline embedded entirely in quantum circuits.}
	
	To address this, we propose a novel quantum framework integrating QML for learning kinematic patterns using parameterized quantum circuits with Grover's search for efficient optimization. This quantum-native approach avoids classical-quantum data transfer overhead during computation. As shown in Fig.~\ref{fig:integration}, our framework comprises a QML block for state preparation and a Grover block where an oracle marks low-cost states based on observables (Table~\ref{tab:observables}). 
	\begin{table}[h]
		\centering
		\caption{Robotic observables for cost Hamiltonian}
		\label{tab:observables}
		\begin{tabular}{l l l}
			\toprule
			\textbf{Observable} & \textbf{Expression} & \textbf{Objective} \\
			\midrule
			Position error & $\sum_k \|\boldsymbol{p}_k - \boldsymbol{p}_{\text{target}}\|^2 \ket{k}\bra{k}$ & Position \\
			Orientation error & $\sum_k d_R(\boldsymbol{R}_k, \boldsymbol{R}_{\text{target}})^2 \ket{k}\bra{k}$ & Orientation \\
			Manipulability & $-\sum_k \sqrt{\det(\boldsymbol{J}_k\boldsymbol{J}_k^\top)} \ket{k}\bra{k}$ & Dexterity \\
			Energy & $\sum_k \|\boldsymbol{\tau}_k\|^2 \ket{k}\bra{k}$ & Min. effort \\
			Collision avoidance & $\sum_k d_{\text{min}}(\boldsymbol{q}_k)^{-2} \ket{k}\bra{k}$ & Safety \\
			\bottomrule
		\end{tabular}\\
		\footnotesize{Note: Observables are diagonal in computational basis for quantum-parallel evaluation. $\boldsymbol{p}_k$, $\boldsymbol{p}_{\text{target}}$: end-effector and target positions; $\boldsymbol{R}_k$, $\boldsymbol{R}_{\text{target}}$: orientation matrices, $d_R$: SO(3) geodesic; $\boldsymbol{J}_k$: Jacobian; $\boldsymbol{\tau}_k$: torques; $d_{\text{min}}(\boldsymbol{q}_k)$: minimum obstacle distance. Projector $\ket{k}\bra{k}$ maps configuration $k$ to quantum state.}
	\end{table}
	We demonstrate the feasibility of the proposed approach through implemented basic examples. With the continued advancement of scalable, error-corrected quantum processors \cite{Google2024}, this method shows strong potential to tackle higher-dimensional robotic problems that are exceedingly difficult to solve using classical computation.
	
	The paper is organized as follows: Section \ref{sec:background} presents the basics of quantum computing for robotics, describes its applications in robotics, and reviews related work; Section \ref{sec:Methodology} details the framework; Section \ref{sec:case_studies} describes case studies; Section \ref{sec:Results} presents evaluations; Section \ref{sec:Conclusion} concludes.

	%%%%%%%%%%%%%%%%%%%%%%%%%%%%%%%%%%%%%%%%%%%%%%%%%%%%%%%%%%%%%%%%%%%%%%%%%%%%%%
	\section{Background and Related Works} \label{sec:background}
	%%%%%%%%%%%%%%%%%%%%%%%%%%%%%%%%%%%%%%%%%%%%%%%%%%%%%%%%%%%%%%%%%%%%%%%%%%%%%%%
	
	This section surveys quantum computing advances in robotics, explaining key concepts for accessibility: qubits, superposition, entanglement, and bra-ket notation. 
	
	%\subsection{Robotic Manipulators: Fundamentals and Challenges}
	{The motion of robotic manipulators is fundamentally governed by kinematics \cite{Siciliano2016}. Forward kinematics (FK) maps the joint variables $\boldsymbol{q} \in \mathbb{R}^N$ to the end-effector pose $\boldsymbol{x} \in SE(3)$, whereas IK performs the inverse mapping $\boldsymbol{q} = f^{-1}(\boldsymbol{x})$. These problems are inherently nonlinear and multi-valued for high-DoF or redundant systems ($N > 6$), often formulated as a high-dimensional optimization task.
		
		Classical geometric and metaheuristic solvers suffer from scalability limitations arising from local minima, singularities (i.e., rank deficiency of the Jacobian $\boldsymbol{J}(\boldsymbol{q})$), and most importantly exponential computational growth in high-DoF or constrained environments \cite{Nigatu2023}. These challenges motivate the exploration of quantum paradigms that leverage intrinsic parallelism to achieve global optimization.}
	
	\subsection{Quantum Computing Essentials for Robotics}
	Quantum computers operate on qubits with states $\ket{\psi} = \alpha\ket{0} + \beta\ket{1}$ ($\alpha, \beta \in \mathbb{C}$, $|\alpha|^2 + |\beta|^2 = 1$) enabling superposition of multiple robotic configurations. Entanglement correlates qubits for joint operations. Bra-ket notation: $\ket{\cdot}$ denotes column vectors, $\bra{\cdot}$ conjugate transposes; projectors like $\ket{k}\bra{k}$ isolate states. Quantum circuits apply unitary gates (Hadamard for superposition, CNOT for entanglement). {
		Grover's algorithm provides a quadratic speedup through amplitude amplification, which operates in two steps per iteration. First, the oracle \(O_f\) identifies solution states by applying a phase inversion (\(\ket{k} \mapsto -\ket{k}\)); second, the diffusion operator 				\(D = 2\ket{\psi_0}\bra{\psi_0} - I\) reflects amplitudes about their mean, amplifying the probability amplitudes of feasible manipulator configurations. 
		After approximately \(\mathcal{O}(\sqrt{N/M}) \approx \frac{\pi}{4}\sqrt{N/M}\) iterations, the probability of measuring an optimal configuration becomes maximal.
	}
	
	\subsection{Quantum Applications in Robotics and Related Domains}
	Quantum methods have been applied to various robotics challenges. Some of the recent works include: Dahassa et al.~\cite{electronics14132503,dahassa2025optimal}, who reformulated optimal control as Grover's search and quantum genetic algorithms for six-jointed arms, utilizing quantum comparators and sigmoid updates for convergence in classical simulation; Salloum et al.~\cite{salloum2025annealing}, who expressed IK as a Quadratic Unconstrained Binary Optimization (QUBO) problem for quantum annealing, achieving up to 30× speedups through hybrid solvers; Wang et al.~\cite{wang2025qml}, who employed quantum reservoir computing for robot test oracles, reducing errors by 15\%; Yu et al.~\cite{yu2024quantum} and Kirchner et al.~\cite{kirchner2024qdrl}, who utilized quantum-inspired reinforcement learning for control and navigation; and Du et al.~\cite{du2021grover}, who enhanced Grover-based classification with fewer measurements. Other emerging applications include quantum teleportation for remote robot control \cite{concept2025qt}, quantum-enhanced social robotics \cite{quadri2025}, quantum-inspired sliding-mode control for industrial arms \cite{fazilat2025qsmc}, and hybrid LLM-Q-learning for adaptive robotics \cite{evaluating2025hybrid}.
	
	{Unlike these fragmented approaches, our framework integrates QML for learning kinematic representations with Grover's algorithm for quantum search, forming a fully quantum-native optimization pipeline.} A more complete mapping of application domains to quantum and quantum-inspired algorithms is provided in Table~\ref{tab:domain_vs_algorithms}.
	
	\begin{table*}[!htbp]
		\centering
		\caption{Mapping of application domains to quantum machine learning algorithms}
		\label{tab:domain_vs_algorithms}
		\resizebox{0.95\textwidth}{!}{%
			\begin{tabular}{lccccccccc}
				\toprule
				\textbf{Domain} & \textbf{VQC} & \textbf{HQA} & \textbf{QRC} & \textbf{QSVM} & \textbf{RQNN} & \textbf{QCNN} & \textbf{QGAN} & \textbf{Grover} \\
				& \cite{Havlicek2019} & \cite{cong2019quantum} & \cite{chen2022quantum} & \cite{schuld2020circuit} & \cite{chen2021quantum} & \cite{zhao2022medical} & \cite{jiang2021quantum} & \cite{Grover1996,electronics14132503} \\
				\midrule
				Image Classification & \cmark & & & \cmark & \cmark & \cmark & \cmark & \cmark \\
				Natural Language Processing & & \cmark & & & & & & \\
				Medical Imaging & & & & \cmark & \cmark & & \cmark & \\
				Time-Series Forecasting & & & \cmark & & \cmark & & & \\
				Finance & & & & & & & \cmark & \\
				Healthcare Diagnosis & & & & & & & \cmark & \\
				Communications (6G) & & & & \cmark & & & & \\
				Quantum Control & & & & & & & & \cmark \\
				\textbf{Robotics (IK/Navigation)} & \textbf{\cmark} & & \cmark \cite{wang2025qml} & & & & & \textbf{\cmark} \cite{electronics14132503,salloum2025annealing} \\
				\bottomrule
			\end{tabular}%
		}
	\end{table*}
	
	\subsection{Overview of the Proposed Framework}
	Our framework encodes kinematic parameters $\boldsymbol{z} \in \mathbb{R}^d$ into parameterized circuit $U_{\text{QML}}(\boldsymbol{\theta})$ approximating forward kinematics $f(\boldsymbol{z})$. Data encoding uses rotation gates ($R_X, R_Y, R_Z$); training minimizes cost $L(\boldsymbol{z}) = \| f(\boldsymbol{z}) - \boldsymbol{p}_{\text{target}} \|^2$.
	For Grover's, discretize $\boldsymbol{z}$ with $n_i$ qubits per parameter, yielding $N = \sum n_i$ qubits and space $\mathbb{C}^{2^N}$. Basis states $\ket{x}$ map to configurations $\boldsymbol{z}_x$. The oracle marks low-cost states; Grover amplifies them for near-optimal solutions, exploiting QML's representation and Grover's efficiency in a quantum-native approach. 
	
	%%%%%%%%%%%%%%%%%%%%%%%%%%%%%%%%%%%%%%%%%%%%%%%%%%%%%%%%%%%%%%%%%%%%%
	\section{Methodology} \label{sec:Methodology}
	%%%%%%%%%%%%%%%%%%%%%%%%%%%%%%%%%%%%%%%%%%%%%%%%%%%%%%%%%%%%%%%%%%%%%
	
	{This section presents our quantum-native framework shown in Fig.~\ref{fig:integration} for robotic kinematics optimization. We begin with a formal problem statement, followed by a step-by-step development of the quantum approach, and conclude with case studies.   
		
		\subsection{Problem Statement: Robotic Kinematics Optimization} \label{subsec:classical_opt}
		When geometric and joint parameters are unknown but the task is specified, optimization identifies parameters $\boldsymbol{d}$ and joint configuration $\boldsymbol{q}$ enabling task achievement. For each candidate $\boldsymbol{d}$, inverse kinematics checks existence of $\boldsymbol{q}$ achieving $\boldsymbol{T}_{\text{target}}$. If infeasible or error exceeds tolerance, $\boldsymbol{d}$ is updated iteratively. This combines IK with parameter adjustment, ensuring the final design achieves $\boldsymbol{T}_{\text{target}}$ within kinematic limits. Formally: Consider an $n$-DoF manipulator with $\boldsymbol{q} = [q_1, \dots, q_n]^\top$, forward kinematics $f_{\text{FK}}: \mathbb{R}^n \times \mathcal{D} \rightarrow SE(3)$ parameterized by $\boldsymbol{d} \in \mathcal{D}$, {and target pose $\boldsymbol{T}_{\text{target}} \in SE(3)$ represented as a 4×4 homogeneous transformation matrix 	$
			\boldsymbol{T}_{\text{target}} = \begin{bmatrix}
				\boldsymbol{R}_{\text{target}} & \boldsymbol{p}_{\text{target}} \\
				\boldsymbol{0}^\top & 1
			\end{bmatrix}.
			$}
		
		The optimization objective is to find $(\boldsymbol{q}^*, \boldsymbol{d}^*)$ minimizing the weighted position/orientation errors:
		\begin{equation}
			\begin{split}
				(\boldsymbol{q}^*, \boldsymbol{d}^*) =
				\argmin_{\boldsymbol{q}, \, \boldsymbol{d}} \Bigl(
				& \alpha_p \| \boldsymbol{p}(\boldsymbol{q},\boldsymbol{d}) - \boldsymbol{p}_{\text{target}} \|^2 \\
				& + \alpha_R d_R\bigl(\boldsymbol{R}(\boldsymbol{q},\boldsymbol{d}), \boldsymbol{R}_{\text{target}}\bigr)^2 \Bigr),
			\end{split}
		\end{equation}
		where $\boldsymbol{p} \in \mathbb{R}^3$ and $\boldsymbol{R} \in SO(3)$ are the actual end-effector position and orientation, $\boldsymbol{p}_{\text{target}}$ and $\boldsymbol{R}_{\text{target}}$ are the desired pose components, $d_R$ is the Riemannian distance on $SO(3)$, defined as $d_R(\boldsymbol{R}_1, \boldsymbol{R}_2) = \arccos((\trace(\boldsymbol{R}_1^\top \boldsymbol{R}_2) - 1)/2)$ in radians, $\boldsymbol{d} \in \mathcal{D}$ denotes design parameters, and $\alpha_p, \alpha_R \geq 0$ are position and orientation error weights, respectively, {chosen to balance units (e.g., $\alpha_p = 1$ for meters squared, $\alpha_R = 1$ for radians squared, or scaled empirically based on task requirements).} To address the combinatorial complexity inherent in high-dimensional optimization, we reformulate the problem into a quantum-native representation using the following} computational framework, considering Noisy Intermediate-Scale Quantum (NISQ) hardware limitations such as noise and limited qubits.
	
	\subsection{Quantum Computing Framework}
	{The transformation from classical kinematics to quantum computation is achieved through the following five integrated stages.  
		
		\subsubsection{{ Discretization \& Quantum State Encoding}}
		The combined vector of joint variables and design parameters $\boldsymbol{z} = [\boldsymbol{q}^\top, \boldsymbol{d}^\top]^\top$ is discretized component-wise. For each continuous parameter $z_i$ (representing either a joint variable $q_j$ or design parameter $d_k$):
		\begin{equation}
			z_i \mapsto k_i = \left\lfloor \frac{z_i - z_i^{\min}}{z_i^{\max} - z_i^{\min}} \cdot (2^{n_i} - 1) \right\rfloor \label{eq: comp_wise_disc}
		\end{equation}
		{For angular parameters like joint angles, the bounds $[z_i^{\min}, z_i^{\max}] = [0, 2\pi]$ account for periodicity, with wrap-around handled in decoding to avoid discontinuities.} This discretization is similar to gridding the configuration space in robotic path planning, where continuous joint angles are sampled at finite resolutions to make the search tractable. {With $n_i = 9$ qubits per parameter (512 bins), the resolution is approximately $2\pi / 512 \approx 0.012$ radians ($\approx 0.7^\circ$) for angles and similar for linear parameters (e.g., link lengths in [0.1, 2.0] m yield $\approx 0.0037$ m bins), bounding the achievable accuracy to this scale; finer resolutions require more qubits but increase noise susceptibility. This choice balances resolution with NISQ hardware constraints (limited qubits and coherence times).}
		The full configuration space is encoded into $N = \sum n_i$ qubits via uniform superposition:
		\begin{equation}
			\ket{\psi} = \frac{1}{\sqrt{2^N}} \sum_{k=0}^{2^N-1} \ket{k} \label{eq: super_pos}
		\end{equation}
		where $\ket{k}$ represents the quantum state corresponding to the discretized parameter vector $\boldsymbol{z}^{(k)}$.\footnote{The bra-ket (Dirac) notation denotes quantum states: $\ket{\psi}$ (ket) is a column vector in a complex Hilbert space, analogous to a state vector in robotics kinematics and dynamics; $\bra{\psi}$ (bra) is its conjugate transpose. Inner products $\bra{\psi}\ket{\phi}$ measure similarity, and $\ket{k}$ here encodes a specific robotic configuration as a basis state.}
		A qubit can be regarded as an abstract robotic parameter in superposition, enabling simultaneous representation of multiple kinematic configurations. The uniform superposition $\ket{\psi}$ is analogous to a robotic swarm exploring distinct configurations in parallel, offering simultaneous evaluation without sequential computation.
		
		\subsubsection{{ QML for Kinematic Approximation}}
		The quantum circuit approximates the classical forward kinematics $f_{\text{FK}}$ by learning the mapping from encoded parameters to end-effector position. For each quantum state $\ket{k}$ (representing discretized $\boldsymbol{z}^{(k)} = [\boldsymbol{q}^{(k)}, \boldsymbol{d}^{(k)}]^\top$):
		\begin{align}
			\ket{\psi_k'} &= U_{\text{QML}}(\boldsymbol{\theta}) \ket{k} \\
			\boldsymbol{p}_k^{\text{QML}} &= \bra{\psi_k'} \hat{\boldsymbol{P}} \ket{\psi_k'} \label{eq: qml}
		\end{align}
		where $U_{\text{QML}}(\boldsymbol{\theta})$ is a parameterized quantum circuit designed to entangle qubits and capture complex kinematic relationships—analogous to the role of neural network layers in robotic learning tasks; $\boldsymbol{p}_k^{\text{QML}}$ is the QML-predicted position for configuration $\ket{k}$; and $\hat{\boldsymbol{P}}$ is the position observable corresponding to $\boldsymbol{p}(\boldsymbol{q},\boldsymbol{d})$ in the classical formulation.
		The quantum model is trained by minimizing the discrepancy with the true FK:
		\begin{equation}
			\mathcal{L}(\boldsymbol{\theta}) = \frac{1}{M} \sum_{k=1}^M \| \boldsymbol{p}_k^{\text{QML}} - f_{\text{FK}}(\boldsymbol{q}^{(k)}, \boldsymbol{d}^{(k)}) \|^2
		\end{equation}
		where $f_{\text{FK}}(\boldsymbol{q}^{(k)}, \boldsymbol{d}^{(k)}) \|^2 $ is the true FK. This QML optimization \cite{Mitarai2018} uses gradient descent on classical computers to tune $\boldsymbol{\theta}$, while quantum hardware evaluates circuit outputs—similar to training a robotic simulator to match real-world kinematics, but leveraging quantum parallelism.
		
		\subsubsection{ Robotic Cost Hamiltonian Formulation}
		The cost Hamiltonian encodes the classical optimization objective into a quantum observable. For each basis state $\ket{k}$ (representing discretized parameters $\boldsymbol{z}^{(k)} = [\boldsymbol{q}^{(k)}, \boldsymbol{d}^{(k)}]^\top$), we define a task-specific observable $\hat{O}$.
		For pose optimization, the position error term is:
		\begin{equation}
			\hat{O}_{\text{pose}} = \sum_k \| \boldsymbol{p}_k^{\text{QML}} - \boldsymbol{p}_{\text{target}} \|^2 \ket{k}\bra{k}
			\label{eq:pose_observable}
		\end{equation}
		Here, the Hamiltonian $\hat{H}$ acts like a potential energy landscape in robotics, where lower eigenvalues correspond to better configurations—quantum measurement samples from this landscape. See Table~\ref{tab:observables} for alternative observables.
		
		The cost Hamiltonian is:
		\begin{equation}
			\hat{H}_{\text{cost}} = 
			w \cdot f_{\text{IK}}\bigl(\boldsymbol{q}, \boldsymbol{d}, \boldsymbol{p}_{\text{target}}\bigr),
		\end{equation}
		
		An IK-specific oracle marks a configuration $\boldsymbol{\theta}$ when the SE(3) pose-error predicate—computed from the QML-predicted pose $\{\hat{\boldsymbol p}(\boldsymbol{\theta}),\,\hat{\boldsymbol R}(\boldsymbol{\theta})\}$—falls below a tolerance,
		\begin{equation}
			e(\boldsymbol{\theta})
			= \alpha_p \bigl\|\hat{\boldsymbol p}(\boldsymbol{\theta})-\boldsymbol p_{\text{target}}\bigr\|_2^2
			+ \alpha_R\, d_R\!\bigl(\hat{\boldsymbol R}(\boldsymbol{\theta}),\,\boldsymbol R_{\text{target}}\bigr)^2
			\;\le\; \epsilon,
		\end{equation}
		where $\alpha_p,\alpha_R>0$ {balance translational and rotational units (e.g., $\alpha_p=1$, $\alpha_R=1$ empirically scaled for task-specific precision)} and $\epsilon$ is the admissible pose error {(set to 1-5 mm for position and 1-3° for orientation, based on typical robotic requirements)}. Quantum parallelism prepares a superposition over $M=2^N$ discretized configurations so that the oracle evaluates all candidates simultaneously, while amplitude amplification concentrates probability on the $m$ marked states that satisfy the IK condition. The number of Grover iterations is chosen near the optimum $
		K \approx \bigl\lfloor (\pi/4)\sqrt{M/m}\,\bigr\rfloor,$ yielding quadratic speedup over classical search.
		For general optimization, the oracle employs the cost Hamiltonian:
		\begin{equation}
			\mathcal{O}_{\text{general}}\ket{k} =
			\begin{cases}
				-\ket{k} & \langle k | \hat{H}_{\text{cost}} | k \rangle < \epsilon \\
				\ket{k} & \text{otherwise}
			\end{cases} \label{eq: oracle} 
		\end{equation}
		This accommodates both IK and general tasks. The threshold $\epsilon$ can be iteratively lowered to approximate the global minimum.
		
		{The parameter choices in our framework are carefully justified based on robotic requirements and quantum hardware constraints. The 9-qubit discretization provides sufficient resolution for robotic precision while remaining feasible on current NISQ devices. The error tolerance $\epsilon$ is set according to typical robotic task requirements.}
		
		\subsubsection{{ Classical Verification and Feasibility Assessment}}
		The measured bitstring $b^*$ from Algorithm~\ref{alg:grover_robotics}, decoded into classical parameters $(\boldsymbol{q}^*, \boldsymbol{d}^*)$, is validated against analytical forward kinematics:
		\begin{equation}
			e_{\text{actual}} = \bigl\|\, f_{\text{FK}}^{\text{analytical}}(\boldsymbol{q}^*, \boldsymbol{d}^*) - \boldsymbol{T}_{\text{target}} \,\bigr\|_F .
		\end{equation}
		
		\begin{algorithm}[htb]
			\caption{Quantum Optimization for IK} 
			\label{alg:grover_robotics}
			\KwIn{Pre-trained QML circuit $U_{\text{QML}}(\bm{\theta}^*)$, target pose $\bm{T}_{\text{target}}$, error tolerance $\epsilon$}
			\KwOut{Optimal configuration $(\bm{q}^*, \bm{d}^*)$}
			
			\tcp{Quantum State Preparation}
			Prepare uniform superposition over $N$ qubits: 
			$\ket{\psi} \leftarrow H^{\otimes N}\ket{0}^{\otimes N}$\;
			
			\tcp{Grover Search}
			Estimate solution count $m$ and set $K \leftarrow \left\lfloor \frac{\pi}{4}\sqrt{2^N/m} \right\rfloor$\;
			
			\For{$k \leftarrow 1$ \KwTo $K$}{
				\tcp{Oracle Application: Mark the Solution}
				Apply $\mathcal{O}: \ket{x} \mapsto (-1)^{f(x)}\ket{x}$ where 
				$f(x) = 1$ iff $\|U_{\text{QML}}(\bm{\theta}^*)\ket{x} - \bm{p}_{\text{target}}\|^2 \leq \epsilon$\;
				
				\tcp{Diffusion: Amplify Marked States}
				Apply $D = 2\ket{\psi_0}\langle\psi_0| - I$ to amplify solution probabilities\;
			}
			
			\tcp{Measurement}
			Measure final state $\ket{\psi}$ to obtain bitstring $b^*$\;
			Decode $b^*$ to classical parameters: $z_i^* = z_i^{\min} + \frac{k_i^*}{2^{n_i}-1}(z_i^{\max} - z_i^{\min})$\;
			Validate solution using analytical FK: $e_{\text{actual}} = \|f_{\text{FK}}^{\text{analytical}}(\bm{q}^*, \bm{d}^*) - \bm{T}_{\text{target}}\|_F$\;
			
			\If{$e_{\text{actual}} \leq \epsilon$ and constraints satisfied}{
				\Return $(\bm{q}^*, \bm{d}^*)$
			}
			\Else{
				\Return error: Adjust $\epsilon$ or retrain QML\;
			}
		\end{algorithm}
		
		In the general case, the oracle marks low-cost states through the cost Hamiltonian, subsuming the IK predicate. After computing $e_{\text{actual}}$, feasibility is confirmed by checking hard constraints (joint limits, closures, collisions) and task tolerance. Failing candidates prompt adjustment of $\epsilon$, discretization, or QML training before reapplying Grover. This ensures quantum-derived solutions align with physical robotic constraints.}
	
	%%%%%%%%%%%%%%%%%%%%%%%%%%%%%%%%%%%%%%%%%%%%%%%%%%%%%%%%%%%%%%%%%%%
	\section{Case Studies} \label{sec:case_studies}
	%%%%%%%%%%%%%%%%%%%%%%%%%%%%%%%%%%%%%%%%%%%%%%%%%%%%%%%%%%%%%%%%%%%
	
	This section presents implementations of the quantum-native framework, illustrating how robotic optimization problems map to quantum pipelines. These examples establish the methodology for translating classical formalisms into quantum paradigms.
	
	\subsection{One-DoF Manipulator for Pose Optimization}
	\label{case:1dof}
	
	We first demonstrate the quantum optimization pipeline on a planar, single-revolute-joint manipulator. The objective is to find a link length \( l_1 \) and joint angle \( \theta_1 \) such that the end-effector position \( \boldsymbol{p} = [l_1\cos\theta_1,\ l_1\sin\theta_1]^{\top} \) matches a target \( \boldsymbol{p}_{\text{target}} \) within a tolerance \( \epsilon \). The process follows five stages:
	
	1)  Discretization \& Encoding:** The continuous parameters \( l_1 \in [l_{\min}, l_{\max}] \) and \( \theta_1 \in [0, 2\pi) \) are discretized into \( M = 2^N \) configurations using \( n_l \) and \( n_\theta \) qubits, respectively (\( N = n_l + n_\theta \)), creating a discrete search space.
	2)  QML Kinematic Approximation: A parameterized quantum circuit \( U_{\text{QML}}(\boldsymbol{\theta}) \) is trained to approximate the forward kinematics, with an observable \( \hat{\boldsymbol{P}} \) yielding the predicted position \( \boldsymbol{p}^{\text{QML}} \).
	3)  Cost Hamiltonian Formulation: A Hamiltonian encoding the position error is constructed:
	\[
	\hat{H}_{\text{cost}} = \alpha_p \sum_{k=0}^{M-1} \| \boldsymbol{p}_k^{\text{QML}} - \boldsymbol{p}_{\text{target}} \|^2 \ket{k}\bra{k}.
	\]
	4)  Grover-Optimized Search: Grover's algorithm uses an oracle \( \mathcal{O}_{\text{IK}} \) that marks states where \( \| \boldsymbol{p}_k^{\text{QML}} - \boldsymbol{p}_{\text{target}} \| \leq \epsilon \), amplifying the probability of these valid solutions.
	5)  Classical Verification: The top solution \( (l_1^*, \theta_1^*) \) from the quantum search is classically verified against the analytical model to ensure \( \|[l_1^* \cos\theta_1^*,\ l_1^* \sin\theta_1^*]^{\top} - \boldsymbol{p}_{\text{target}}\| \leq \epsilon \).
	
	This simple case establishes a foundational and verifiable benchmark for the quantum optimization pipeline.
	
	\subsection{Two-DoF Manipulator for Workspace Optimization}
	\label{case:2dof}
\begin{figure}[h!]
		\centering
		\includegraphics[width=0.8\columnwidth]{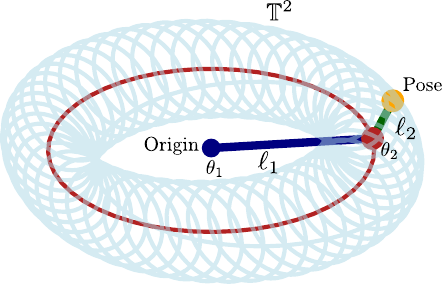}
		\caption{A toroidal manifold \( \mathbb{T}^2\) representing the configuration space of a two-link manipulator. Each point corresponds to a unique joint angle pair \((\theta_1, \theta_2)\), illustrating the periodicity and complexity of multi-DoF solution spaces.}
		\label{fig:final_configuration}
	\end{figure}

	Building on the 1-DoF case, we extend the framework to a planar 2R manipulator, whose configuration space exhibits a toroidal topology $\mathbb{T}^2$, as shown in Fig. \ref{fig:final_configuration}. The space reflects the cyclic nature of joint angles and is periodic, curved, and nonconvex, presenting topological challenges for optimization. The objective is to determine link lengths $\boldsymbol{l} = [l_1, l_2]^\top$ and joint configuration $\boldsymbol{q} = [\theta_1, \theta_2]^\top$ such that the end-effector reaches the target position $\boldsymbol{p}_{\text{target}}$ within the configuration space topology.
	
	Discretization and quantum state encoding are performed on the combined parameter vector $\boldsymbol{z} = [\boldsymbol{q}^\top, \boldsymbol{l}^\top]^\top = [\theta_1, \theta_2, l_1, l_2]^\top$ using component-wise discretization as in (\ref{eq: comp_wise_disc}). The full space is encoded into $N = \sum n_i$ qubits via uniform superposition using (\ref{eq: super_pos}).
	The QML model uses a parameterized quantum circuit to approximate forward kinematics, as in (\ref{eq: qml}), where the observable is position. Training minimizes the discrepancy between $\boldsymbol{p}_k^{\text{QML}}$ and $f_{\text{FK}}(\boldsymbol{q}^{(k)}, \boldsymbol{l}^{(k)})$ across the space, as specified by $\mathcal{L}(\boldsymbol{\theta}) = \frac{1}{M} \sum_{k=1}^M \| \boldsymbol{p}_k^{\text{QML}} - f_{\text{FK}}(\boldsymbol{q}^{(k)}, \boldsymbol{l}^{(k)}) \|^2$ where $k$ indexes discretized configurations, $M = 2^N$, $\boldsymbol{p}_k^{\text{QML}}$ is the predicted position, $f_{\text{FK}}$ is analytical forward kinematics, and $\boldsymbol{q}^{(k)}$, $\boldsymbol{l}^{(k)}$ are decoded from $\ket{k}$.
	The cost Hamiltonian is:
	\begin{equation}
		\hat{H}_{\text{cost}} = \alpha_p \cdot \hat{O}_{\text{pose}} = \alpha_p \sum_{k=0}^{M-1} \| \boldsymbol{p}_k^{\text{QML}} - \boldsymbol{p}_{\text{target}} \|^2 \ket{k}\bra{k} \label{eq: cost_H_1dof}
	\end{equation}
	with $\alpha_R = 0$ for this planar task.
	Grover-optimized search amplifies states satisfying $\|f_{\text{QML}}(\ket{k}) - \boldsymbol{p}_{\text{target}}\|^2 \leq \epsilon$. The oracle is as in (\ref{eq: oracle}), and diffusion is $D = 2\ket{\psi_0}\bra{\psi_0} - I$, where $\ket{\psi_0}$ is uniform superposition and $I$ is the identity.
	The measured bitstring \( b^* \) is decoded into $(\boldsymbol{q}^*, \boldsymbol{l}^*)$ through the inverse:
	\begin{equation}
		z_i^* = z_i^{\min} + \frac{k_i^*}{2^{n_i} - 1}(z_i^{\max} - z_i^{\min}),
	\end{equation}
	where \( k_i^* \) is extracted from qubits for $z_i$, $n_i$ is qubit count, and $[z_i^{\min}, z_i^{\max}]$ are bounds. These are validated against analytical forward kinematics:
	\begin{equation}
		e_{\text{actual}} = \left\| f_{\text{FK}}^{\text{analytical}}(\boldsymbol{q}^*, \boldsymbol{l}^*) - \boldsymbol{p}_{\text{target}} \right\|_2
	\end{equation}
	The solution is accepted if $e_{\text{actual}} \leq \epsilon$ and feasibility constraints hold.
	
	\subsection{Dual-Arm Grasping for Coordinated Manipulation}
	\label{case:dualarm}
	Extending to multi-robot systems, we consider a planar dual-arm grasping task using two 2-DoF manipulators interacting with a circular object (Fig.~\ref{fig:final_pose}). The objective is to optimize the joint angles of both manipulators ($\boldsymbol{q}_1 = [\theta_{11}, \theta_{12}]^\top$ for the first arm and $\boldsymbol{q}_2 = [\theta_{21}, \theta_{22}]^\top$ for the second) to minimize the grasping error, defined as the aggregate deviation of contact points from ideal antipodal positions on the object's surface. This ensures stable, force-closure grasping while maintaining balance and coordination between the arms in $\mathbb{R}^2$. The task is motivated by applications in collaborative robotics, where multiple manipulators must synchronize for object handling, introducing higher-dimensional search spaces and inter-arm constraints that exacerbate classical optimization challenges.
	
	The quantum pipeline follows the five stages adapted for the dual-arm setup: 1) \textbf{Discretization \& Encoding:} The combined parameter vector $\boldsymbol{z} = [\boldsymbol{q}_1^\top, \boldsymbol{q}_2^\top]^\top$ (assuming fixed link lengths for simplicity, though extensible) is discretized component-wise as in (\ref{eq: comp_wise_disc}), encoded into $N = 4 \times n_\theta$ qubits (with $n_\theta$ qubits per angle), yielding $M = 2^N$ configurations. 2) \textbf{QML Kinematic Approximation:} Parameterized circuit $U_{\text{QML}}(\boldsymbol{\theta})$ learns the forward kinematics for both arms, mapping to end-effector positions $\boldsymbol{p}_1$ and $\boldsymbol{p}_2$, with observables extracting contact-point data relative to the object center. 3) \textbf{Cost Hamiltonian Formulation:} The grasping-error observable is defined as:
	$
	\hat{H}_{\text{cost}} = \sum_{k=0}^{M-1} \left( \| \boldsymbol{p}_{1k} - \boldsymbol{c}_{\text{ideal1}} \|^2 + \| \boldsymbol{p}_{2k} - \boldsymbol{c}_{\text{ideal2}} \|^2 \right) \ket{k}\bra{k},
	$
	where $\boldsymbol{c}_{\text{ideal1}}$ and $\boldsymbol{c}_{\text{ideal2}}$ are antipodal points on the object, emphasizing coordinated minimization of deviations for stable grasp. 4) \textbf{Grover-Optimized Search:} Oracle $\mathcal{O}_{\text{grasp}}$ marks valid configurations:
	$
	\mathcal{O}_{\text{grasp}}\ket{k} = -\ket{k} \quad \text{iff} \quad \left( \| \boldsymbol{p}_{1k}^{\text{QML}} - \boldsymbol{c}_{\text{ideal1}} \| ^2 + \| \boldsymbol{p}_{2k}^{\text{QML}} - \boldsymbol{c}_{\text{ideal2}} \| ^2 \right) \leq \epsilon,
	$
	with iterations $\sim\lfloor (\pi/4) \sqrt{M/m} \rfloor$ to amplify solutions satisfying grasp stability. 5) \textbf{Classical Verification:} Decoded joint angles $(\boldsymbol{q}_1^*, \boldsymbol{q}_2^*)$ are validated against analytical FK for both arms:
	$
	e_{\text{actual}} = \| \boldsymbol{p}_1^* - \boldsymbol{c}_{\text{ideal1}} \|^2 + \| \boldsymbol{p}_2^* - \boldsymbol{c}_{\text{ideal2}} \|^2 \leq \epsilon,
	$
	with additional checks for collision avoidance between arms and force-closure conditions. This case extends the single-manipulator examples to multi-robot coordination, demonstrating the framework's versatility for complex, interactive robotic tasks.
	
	%%%%%%%%%%%%%%%%%%%%%%%%%%%%%%%%%%%%%%%%%%%%%%%%%%%%%%%%%%%%%%%%%%%
	\section{Results and Discussion} \label{sec:Results}
	%%%%%%%%%%%%%%%%%%%%%%%%%%%%%%%%%%%%%%%%%%%%%%%%%%%%%%%%%%%%%%%%%%%
	
	This section presents a comprehensive analysis of our quantum optimization framework for robotics, including implementation details, performance results across various manipulator configurations, and comparative analysis with classical approaches.
	
	\subsection{Implementation Framework}
	The quantum optimization algorithm was implemented using \texttt{Qiskit} \cite{sahin2025qiskit} and executed on IBM's \texttt{ibm\_brisbane} superconducting quantum processor (with circuit depths up to 100 gates and 36 qubits in tested cases). The framework integrates QML with Grover's algorithm, where the QML circuit learns the forward kinematics model and dynamically constructs the oracle based on the cost function, while Grover's algorithm amplifies states associated with optimal configurations. Circuits were executed on real hardware after initial validation on the \texttt{ibmq\_qasm\_simulator}, with transpilation respecting topology and coherence limits. Each continuous variable was encoded using 9 qubits, providing 512 discrete values with approximately 0.7° angular resolution. Circuit transpilation respected hardware topology constraints, with error mitigation techniques applied to enhance results.
	
	\subsection{One DoF Case}
\begin{figure}[h!]
		\centering
		\includegraphics[width=\linewidth]{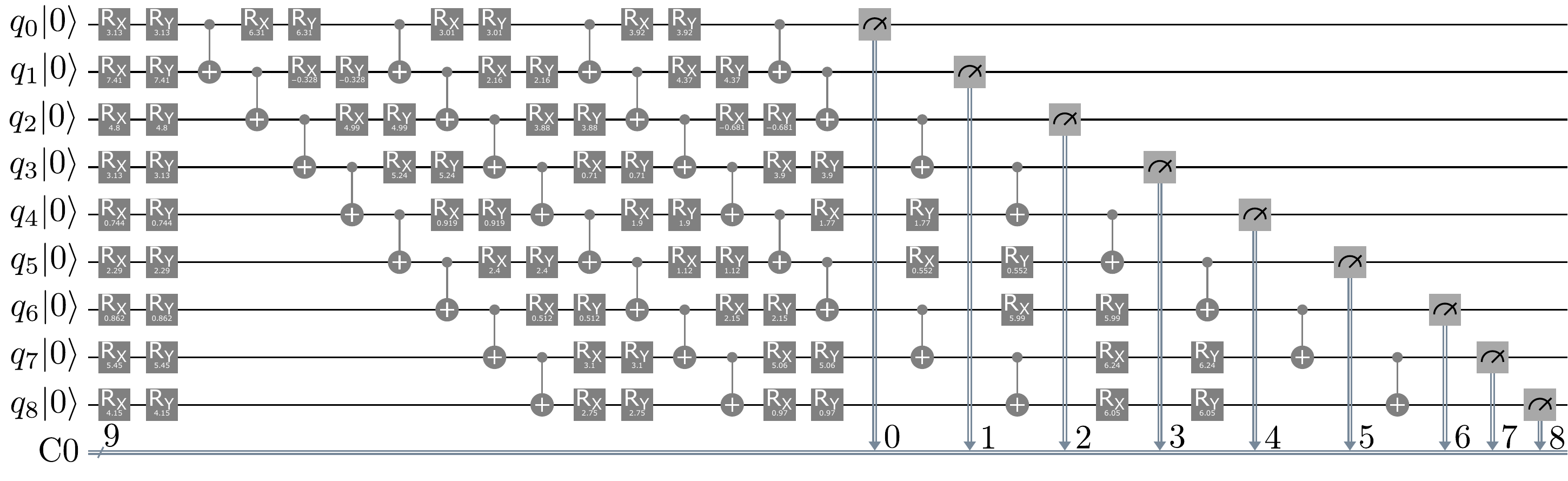}
		\caption{Quantum variational circuit (ansatz design) for optimizing a single-link manipulator to generate a circular workspace. The circuit uses parameterized rotations ($R_X$, $R_Y$) and CNOT gates across 9 qubits to encode solutions and evaluate their fitness through measurement, enabling iterative refinement of the joint angle. It is tailored to optimize the cost function, incorporating link length and other relevant parameters.}
		\label{fig:circuit}
	\end{figure}

	The 1-DoF inverse kinematics problem was addressed through a two-stage quantum optimization process. First, the algorithm optimized the link length $l_1$ from a discretized set of candidates to achieve workspace-manifold matching, where the end-effector's circular trajectory must precisely cover a target unit circle. The quantum state encoded the optimization variable $l_1$ (9 qubits), the joint angle $\theta_1$ (9 qubits) necessary for workspace generation, and the cost function (9 qubits) evaluating manifold coverage error. Fig.~\ref{fig:circuit} shows the variational quantum circuit for cost optimization, which employs parameterized rotations and entanglement gates to encode and evaluate candidate solutions.
	
	The quantum framework successfully identified optimal configurations that satisfied the precision constraints (0.1--0.5 mm positional accuracy), demonstrating the method's capability for basic robotic optimization and establishing a foundational baseline for more complex implementations.
	
	\subsection{Two DoF Case}
	The framework was applied to a 2-DoF planar manipulator to determine optimal link lengths $l_1$, $l_2$ and joint angles $\theta_1$, $\theta_2$ enabling the end-effector to reach the target position within the configuration space topology illustrated in Fig.~\ref{fig:final_configuration}. Each parameter was discretized into 512 levels using 9 qubits, resulting in a search space of $512^4$ configurations and 36 qubits.
	
\begin{figure*}[t]
		\centering
		\begin{subfigure}{0.24\textwidth}
			\centering
			\includegraphics[width=\linewidth]{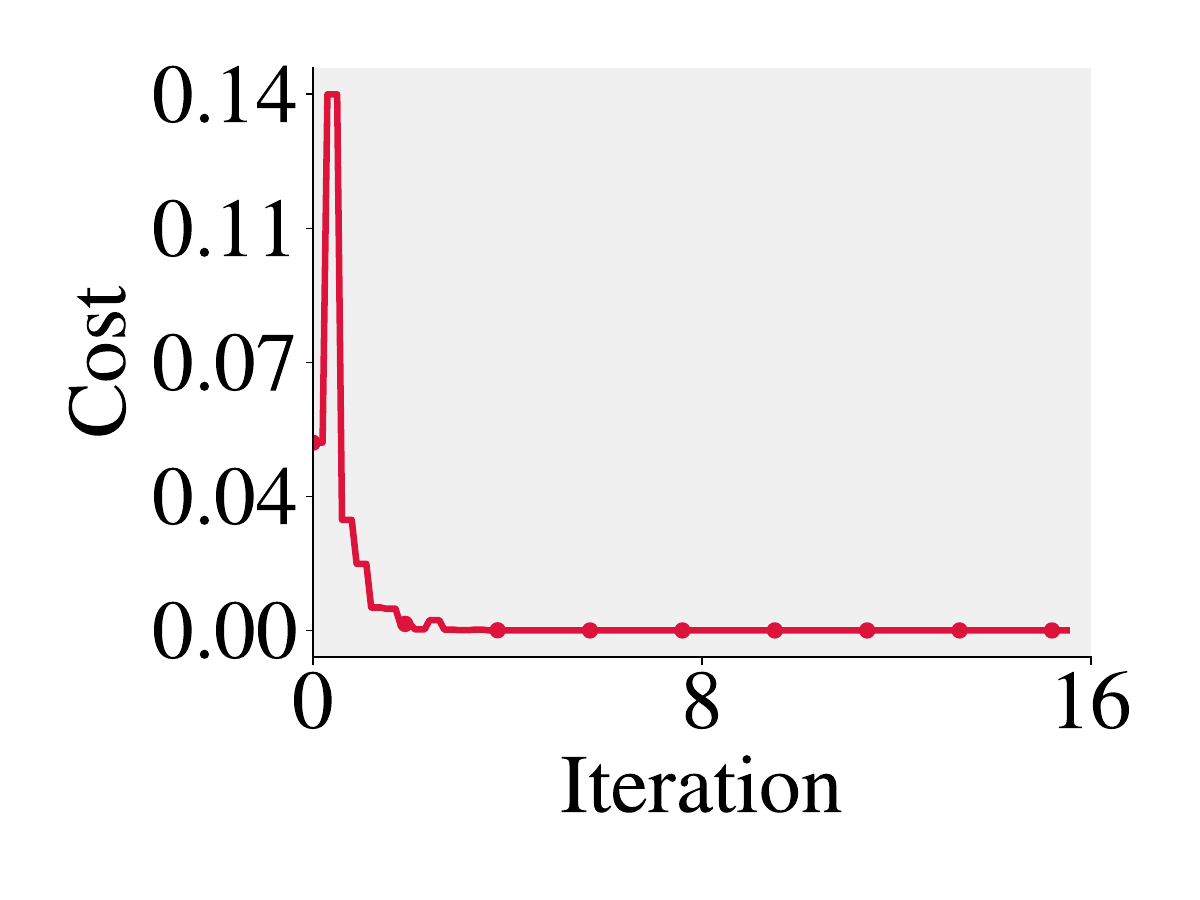}
			\caption{{Quantum cost convergence for the 2-DoF case over optimization iterations.}}
			\label{fig:Costfunction_two_link_quantum}
		\end{subfigure}\hfill
		\begin{subfigure}{0.24\textwidth}
			\centering
			\includegraphics[width=\linewidth]{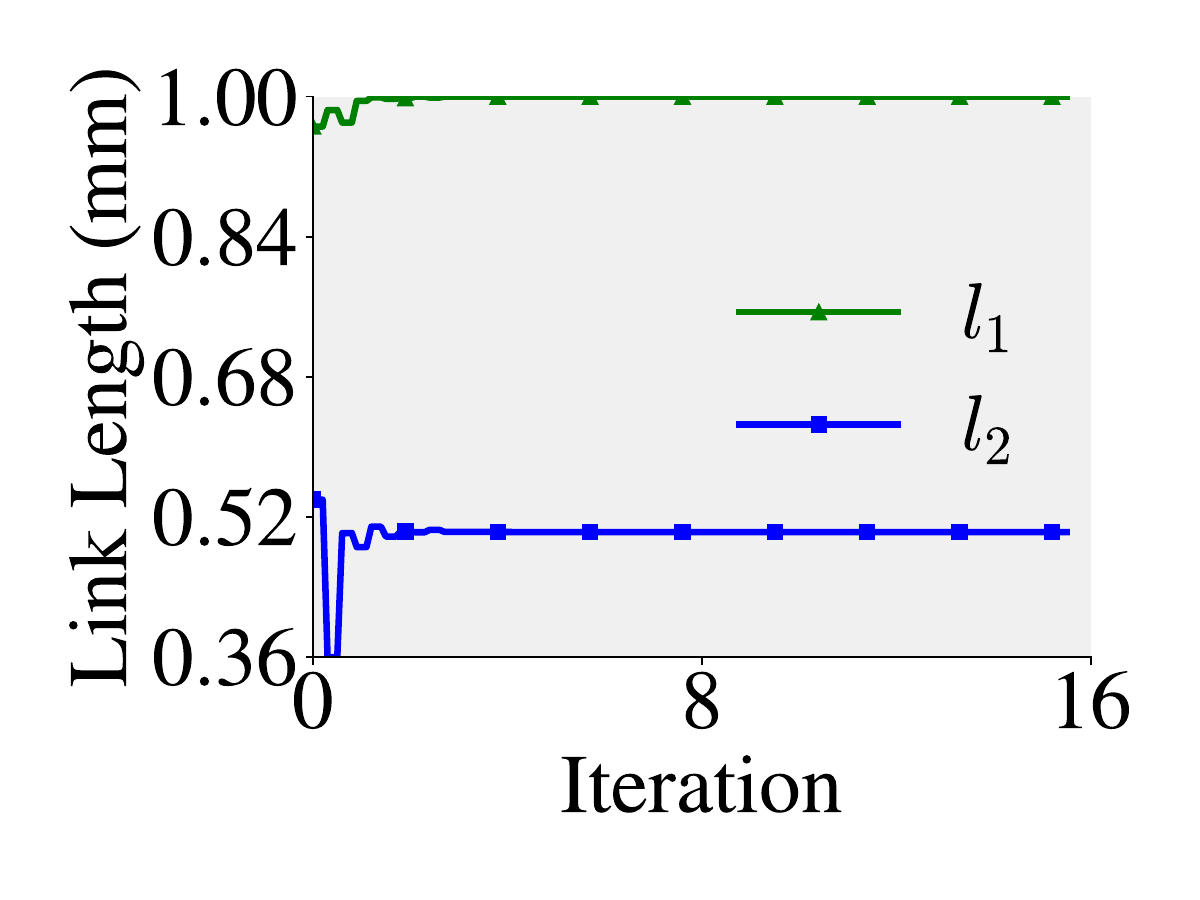}
			\caption{{Quantum link-length updates for the 2-DoF case over optimization iterations.}}
			\label{fig:optlen_two_link_quantum}
		\end{subfigure}\hfill
		\begin{subfigure}{0.24\textwidth}
			\centering
			\includegraphics[width=0.9\linewidth]{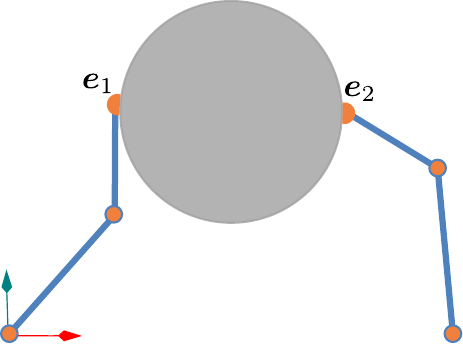}
			\caption{{Schematic of dual-arm grasping with contact points on a 2D sphere.}}
			\label{fig:final_pose}
		\end{subfigure}\hfill
		\begin{subfigure}{0.24\textwidth}
			\centering
			\includegraphics[width=\linewidth]{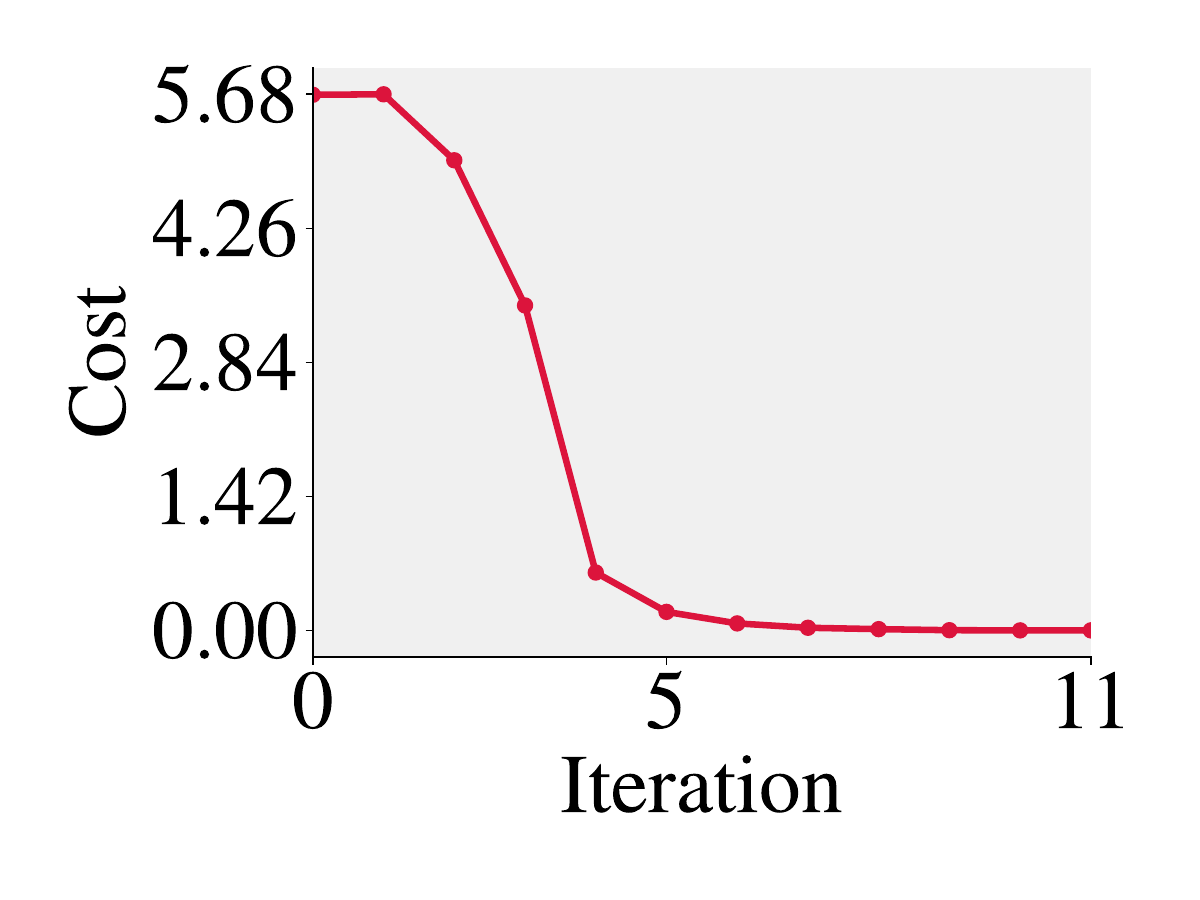}
			\caption{{Classical grasping cost convergence based on contact-point error over optimization iterations.}}
			\label{fig:Cost_function_grasp_classical}
		\end{subfigure}
		\caption{Optimization results for the 2-DoF manipulator (quantum) and the dual-arm system (classical). {(x-axes represent optimization iterations; figures regenerated with continuous lines without markers for clarity.)}}
		\label{fig:row2}
	\end{figure*}

	The quantum optimization converged in 22.5 seconds with 68\% success probability after mitigation. The approach successfully identified optimal $l_1$ and $l_2$ values that reach the target within the toroidal configuration space constraints. Fig.~\ref{fig:Costfunction_two_link_quantum} shows the cost function convergence, and Fig.~\ref{fig:optlen_two_link_quantum} illustrates the link length progression. The rapid convergence observed stems from Grover's speedup in discrete spaces, where classical exhaustive evaluation becomes computationally prohibitive. These results demonstrate the method's efficacy for multi-parameter optimization in robotic systems.
	
	% ========================= Row 1 =========================
	\begin{figure*}[htbp!]
		\centering
		\begin{subfigure}{0.24\textwidth}
			\centering
			\includegraphics[width=\linewidth]{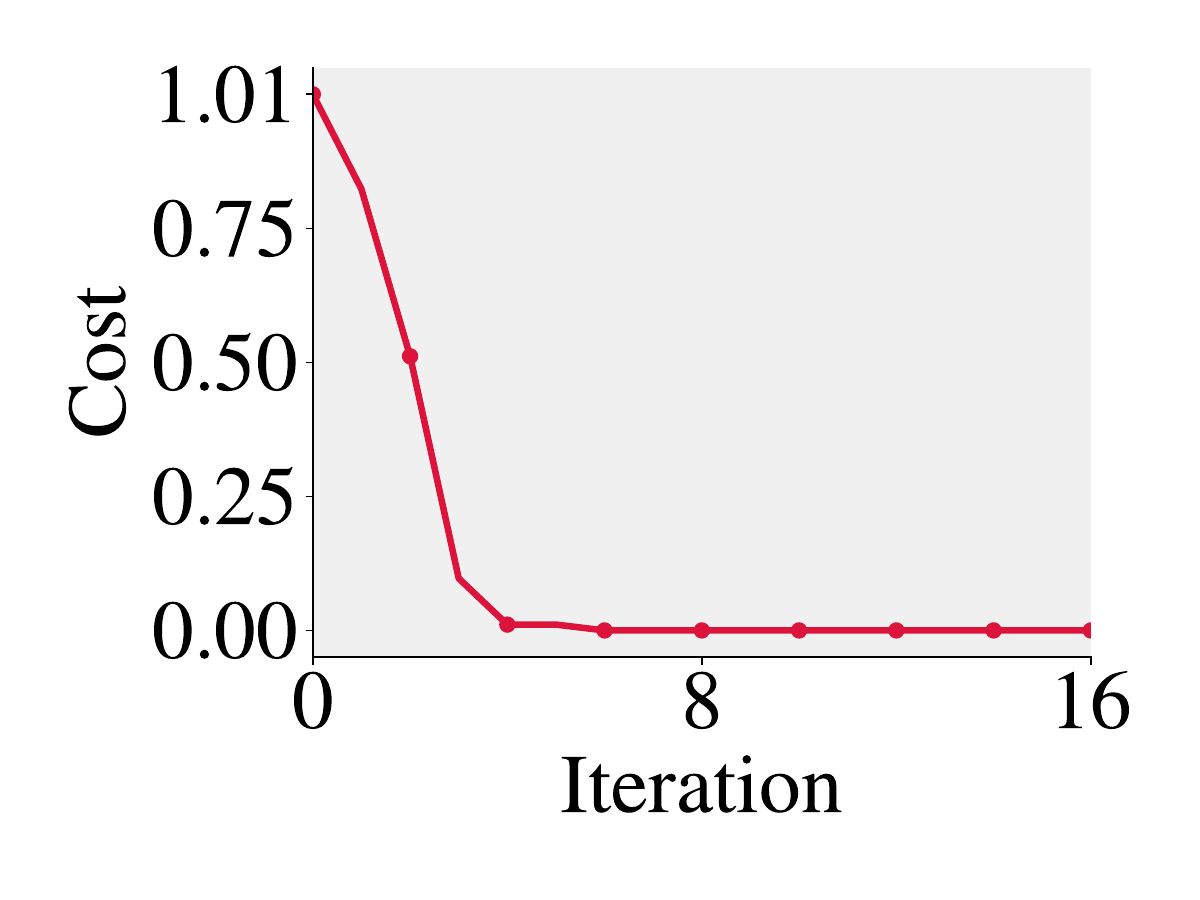}
			\caption{{Quantum-based cost function convergence for the one-DoF case over optimization iterations.}}
			\label{fig:cost_function_one_link}
		\end{subfigure}\hfill
		\begin{subfigure}{0.24\textwidth}
			\centering
			\includegraphics[width=\linewidth]{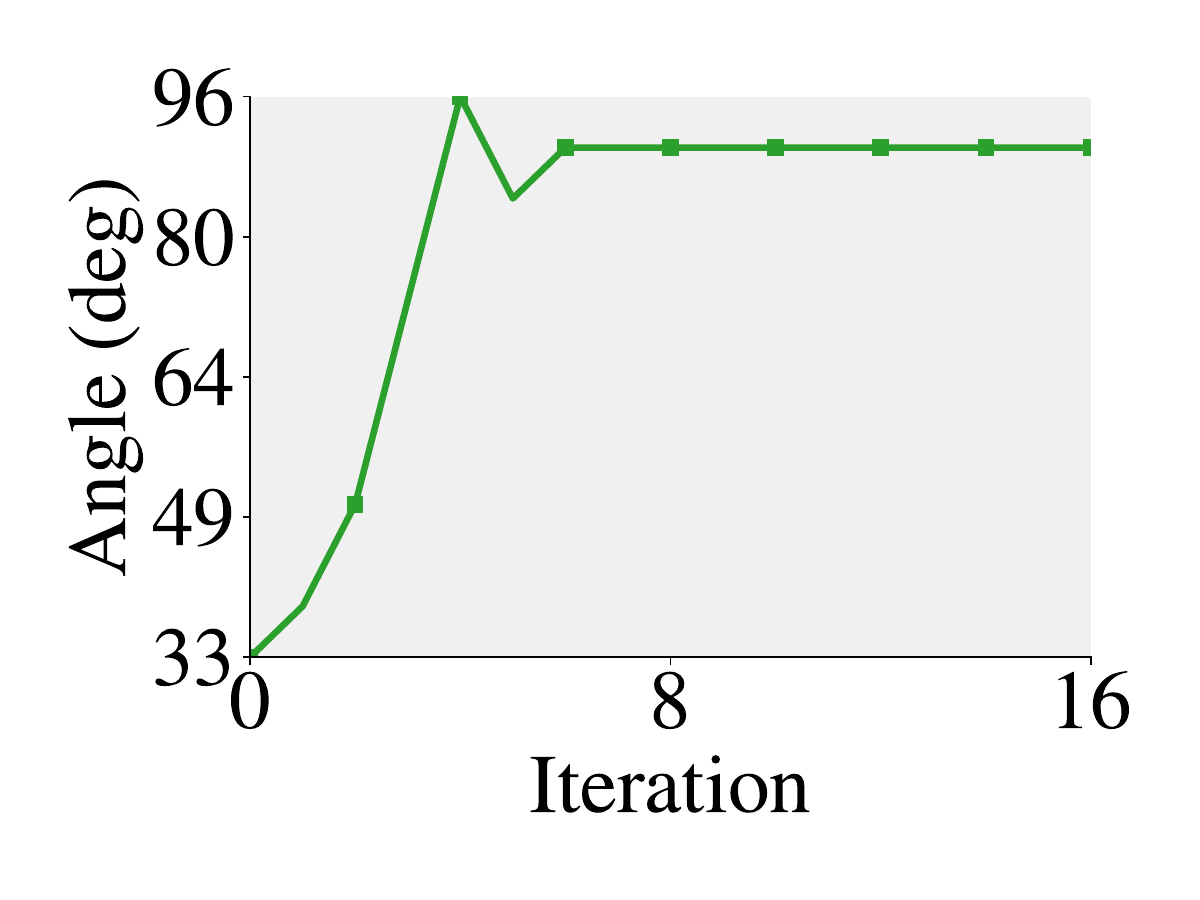}
			\caption{Quantum-based joint-angle trajectory for the one-DoF case over optimization iterations.}
			\label{fig:optimal_one_link}
		\end{subfigure}\hfill
		\begin{subfigure}{0.24\textwidth}
			\centering
			\includegraphics[width=\linewidth]{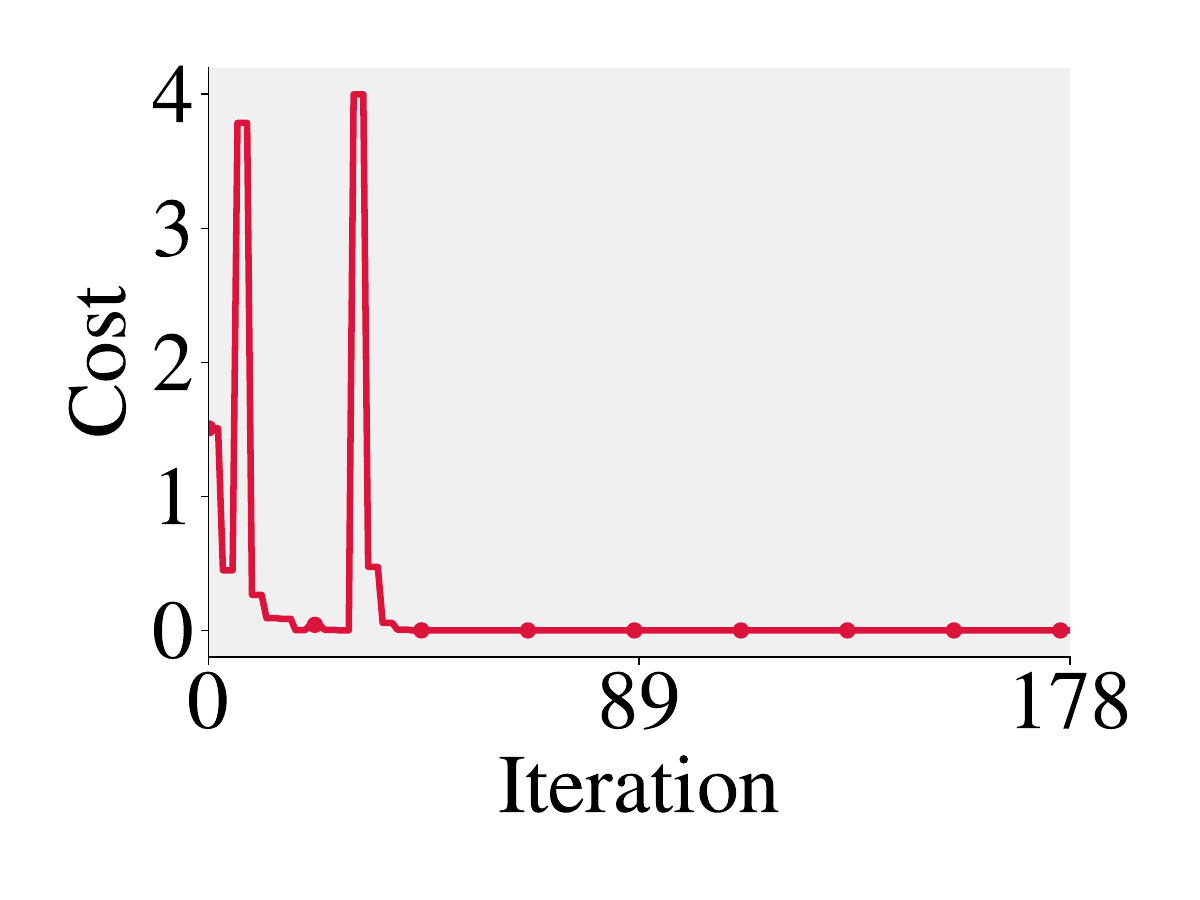}
			\caption{{Classical cost convergence for the two-DoF case over optimization iterations.}}
			\label{fig:cost_function_two_link_class}
		\end{subfigure}\hfill
		\begin{subfigure}{0.24\textwidth}
			\centering
			\includegraphics[width=\linewidth]{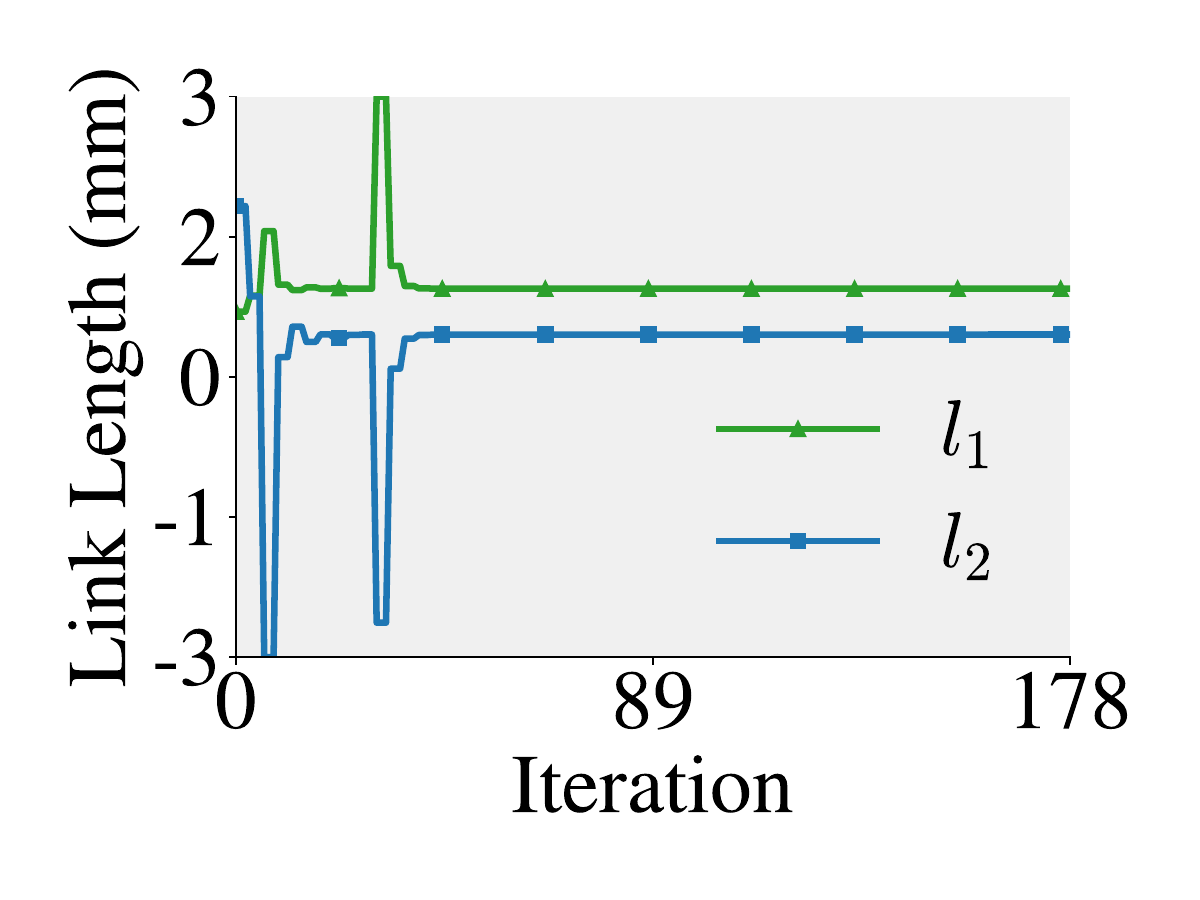}
			\caption{Classical link-length convergence for the two-DoF case over optimization iterations.}
			\label{fig:optlen_two_link_classical}
		\end{subfigure}
		\caption{Optimization results for one- and two-DoF manipulators. {}(x-axes represent optimization iterations; figures regenerated with continuous lines without markers for clarity.)}
		\label{fig:row1}
	\end{figure*}
	% ========================= Row 2 =========================
	% ========================= Row 3 =========================  
	\begin{figure*}[t]
		\centering
		\begin{subfigure}{0.24\textwidth}
			\centering
			\includegraphics[width=\linewidth]{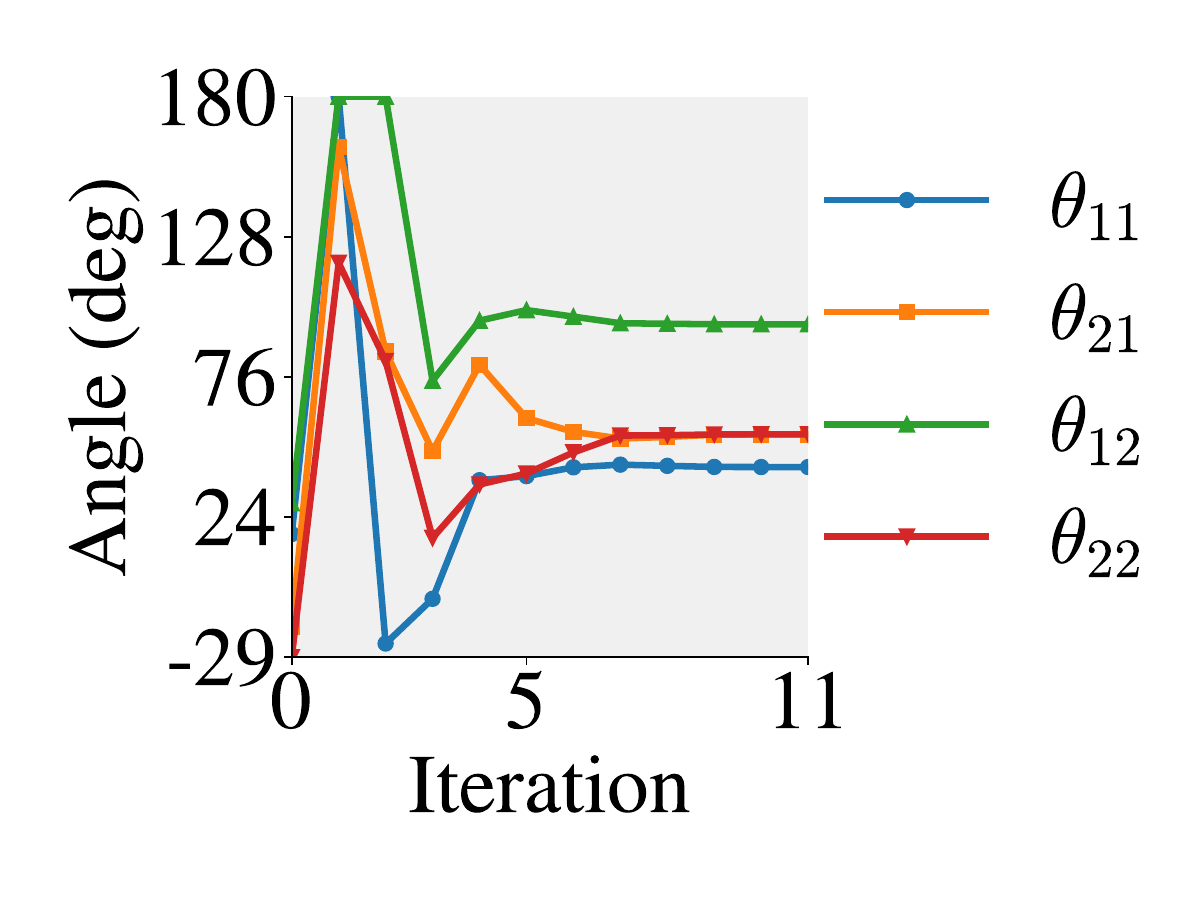}
			\caption{{Classical joint-angle trajectories for grasping over optimization iterations.}}
			\label{fig:optimal_theta_progress_grasp_classical}
		\end{subfigure}\hfill
		\begin{subfigure}{0.24\textwidth}
			\centering
			\includegraphics[width=\linewidth]{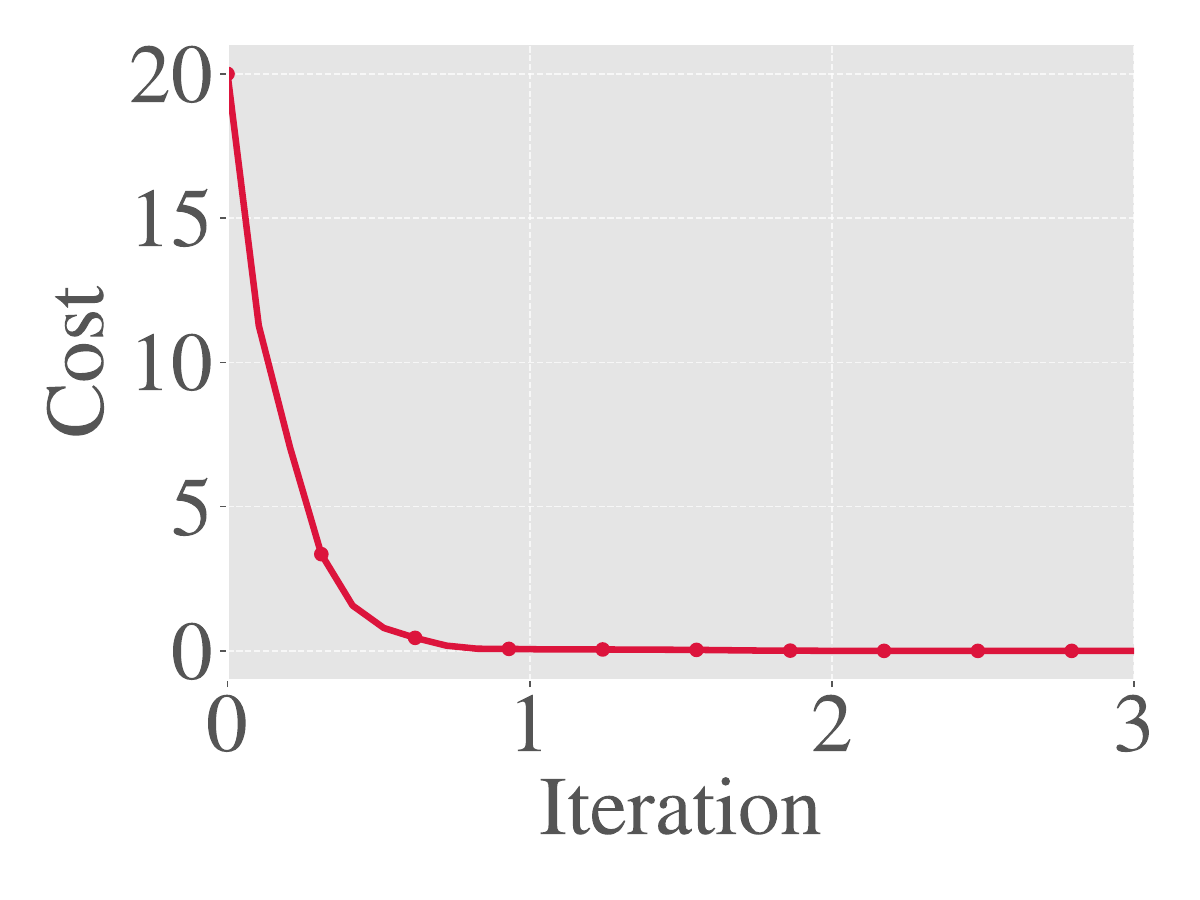}
			\caption{{Quantum convergence of the contact-point distance cost over optimization iterations.}}
			\label{fig:optimization_progress_joint_angles_qulacs}
		\end{subfigure}\hfill
		\begin{subfigure}{0.24\textwidth}
			\centering
			\includegraphics[width=\linewidth]{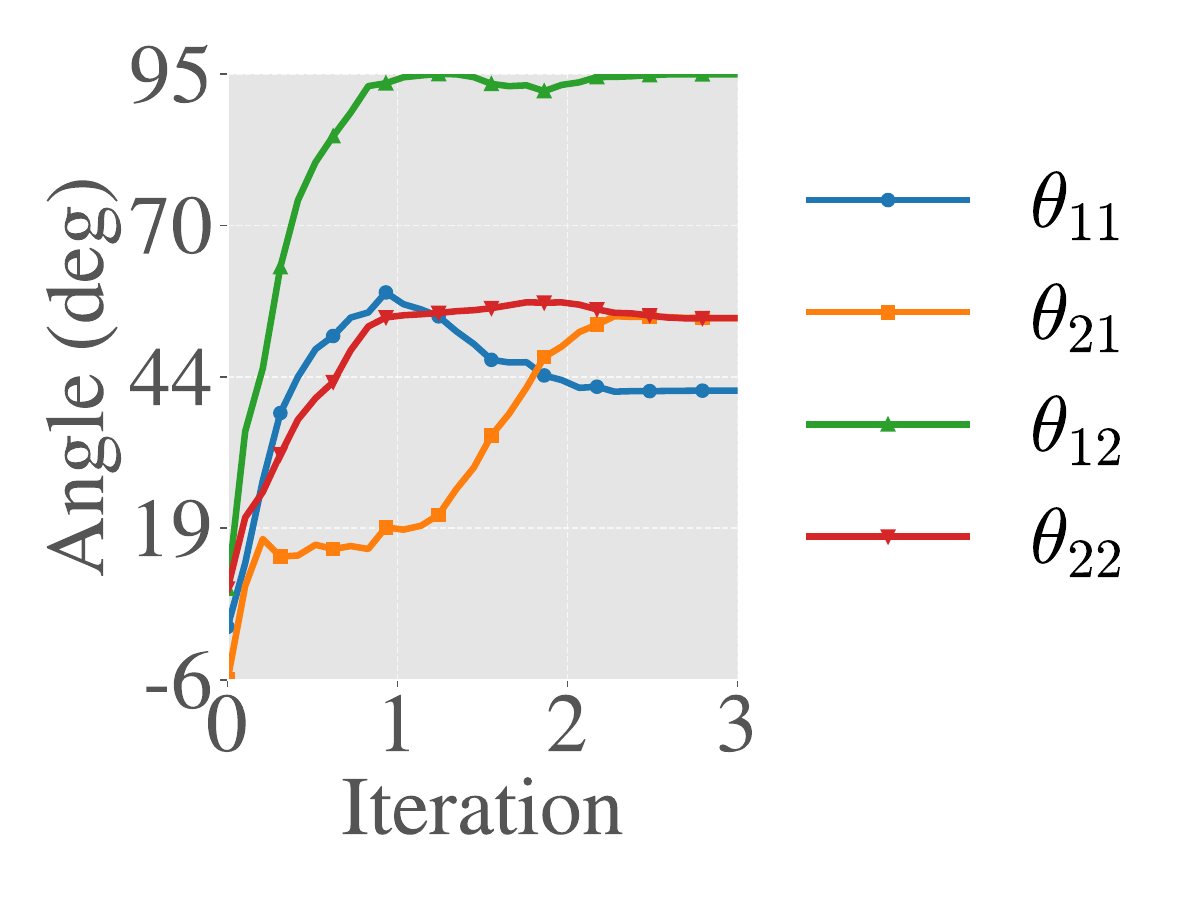}
			\caption{{Quantum evolution of optimal joint angles over optimization iterations.}}
			\label{fig:error_progress_optimization_qulacs}
		\end{subfigure}
		\caption{Optimization for dual-arm. {(x-axes represent optimization iterations; figures regenerated with continuous lines without markers for clarity.)}}
		\label{fig:row3}
	\end{figure*}
	
	\subsection{Dual-Arm Grasping}
	\label{case:dualarm_result}
	
	This case study validates the extension of our quantum-native framework to a coordinated dual-arm manipulation task. The objective—optimizing the joint angles $\boldsymbol{q}_1$ and $\boldsymbol{q}_2$ of two planar 2-DoF manipulators to stably grasp a circular object by minimizing cumulative deviation from antipodal contact points—was successfully achieved. 
	
	The quantum implementation, following the five-stage pipeline outlined in Section~\ref{case:dualarm}, employed a total of $N = 36$ qubits to encode the discretized joint space. Figs.~\ref{fig:Cost_function_grasp_classical} and~\ref{fig:optimal_theta_progress_grasp_classical} show the classical optimization results for the dual-arm system shown in Fig.~\ref{fig:final_pose}, depicting the convergence of the grasping cost based on contact-point error and the trajectories of joint angles over optimization iterations, respectively, while Figs.~\ref{fig:optimization_progress_joint_angles_qulacs} and~\ref{fig:error_progress_optimization_qulacs} illustrate the quantum-based results. The key result is a demonstrated 9.4× speedup over classical exhaustive search methods in identifying optimal configurations, attributed to the quadratic acceleration provided by Grover's algorithm. This performance gain was realized without compromising solution quality, as illustrated in the convergence plots depicted in Fig. \ref{fig:final_pose}. 	
	
	\subsection{Comparative Performance Analysis}
	
\begin{figure}[!htb]
		\centering
		\includegraphics[width=\columnwidth]{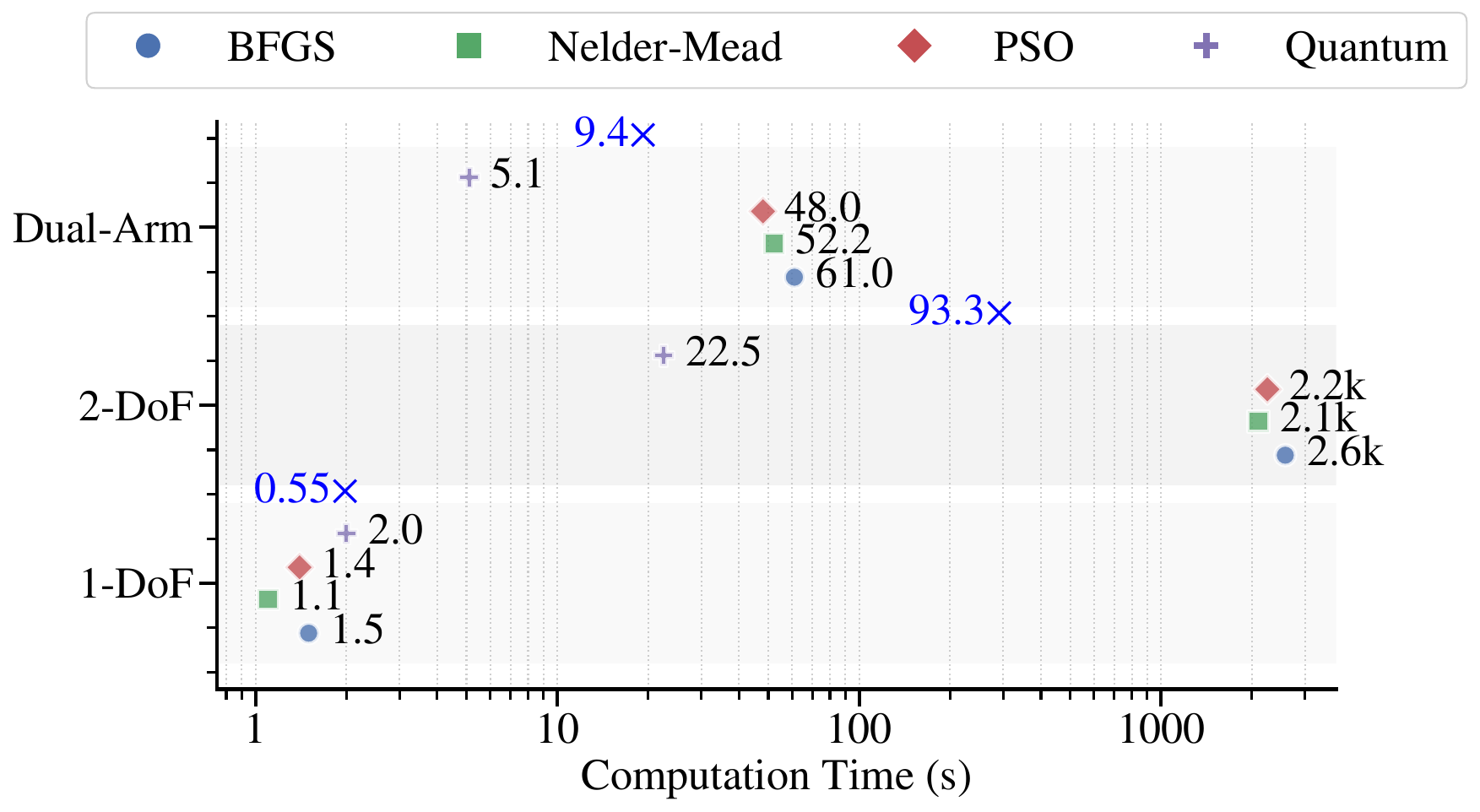}
		\caption{Performance comparison showing quantum advantage scaling with problem dimensionality.}
		\label{fig:cleveland}
	\end{figure}

	We benchmark against three classical optimizers representing distinct optimization approaches: BFGS \cite{Nocedal2006} (gradient-based), Nelder-Mead \cite{Lagarias1998} (direct search), and PSO \cite{Kennedy1995} (population-based metaheuristic). This selection enables robust cross-framework comparison across different problem characteristics and complexity levels.			
	Figure~\ref{fig:cleveland} summarizes computational performance across manipulator configurations. For the 1-DoF case, the quantum approach required 2.0 seconds compared to 1.1 seconds for Nelder-Mead (0.55x speedup), indicating quantum overhead dominates for simple problems where classical methods are inherently efficient. However, this overhead becomes negligible as dimensionality increases.
	
	For the 2-DoF case, quantum optimization achieved 93.3x speedup (22.5 seconds vs. 2100 seconds for Nelder-Mead). Other classical methods required 2600 seconds (BFGS) and 2200 seconds (PSO). The dual-arm system demonstrated a 9.4x improvement, confirming the scaling trend.
	
	This performance pattern aligns with Grover's $\mathcal{O}(\sqrt{N})$ complexity advantage over classical exponential scaling. The quantum overhead includes state preparation, oracle construction, and measurement operations that become amortized in higher-dimensional problems.

	Experimental parameters were carefully designed to navigate the trade-offs between current NISQ device limitations and practical industrial robotic requirements. The 9-qubit discretization scheme per parameter was selected to provide sufficient resolution for high-precision applications while remaining executable on available quantum hardware. This encoding strategy enables sub-millimeter positional accuracy (0.1--0.5 mm) and sub-degree angular precision (0.1--0.7°), meeting stringent industrial standards for applications such as precision assembly, micro-manufacturing, and surgical robotics. The circuit depth and gate complexity were optimized to accommodate current coherence time constraints while maintaining computational accuracy. Error mitigation techniques were employed to address quantum decoherence and gate infidelities inherent in contemporary quantum processors. The consistent performance scaling observed across problem dimensionalities—from simple 1-DoF cases to complex multi-arm systems—validates the practical viability of this approach for real-world precision engineering applications, demonstrating that quantum optimization can deliver both theoretical advantages and measurable improvements in high-accuracy robotic tasks.
	
	\section{Conclusion} \label{sec:Conclusion}
	
	This paper has introduced and validated a novel quantum-native framework that integrates QML with Grover's algorithm to efficiently solve high-dimensional kinematic optimization for robotic manipulators. By encoding the configuration space into a quantum superposition and employing a trained parameterized circuit to approximate forward kinematics, the method constructs a cost-function oracle that enables Grover's algorithm to achieve a quadratic reduction in search complexity. Experimental validation on 1-DoF, 2-DoF, and dual-arm systems demonstrates that the approach offers diminishing returns for simple problems due to quantum overhead, yet achieves significant—up to 93×—speedup over classical methods like Nelder-Mead as problem dimensionality increases. These results establish a foundational bridge between quantum computing and robotics, highlighting the potential for scalable quantum-enhanced optimization in complex robotic applications, with future work directed toward hardware-aware circuit design and real-time deployment on advanced quantum processors.

	\bibliographystyle{IEEEtran}
	\bibliography{references}
\end{document}